\documentclass[letterpaper]{article} 
\usepackage[utf8]{inputenc}
\usepackage{newunicodechar}
\newunicodechar{ }{~} 
\usepackage{aaai2026}  
\usepackage{times}  
\usepackage{helvet}  
\usepackage{courier}  
\usepackage[hyphens]{url}  
\usepackage{graphicx} 
\usepackage{natbib}  
\usepackage{caption} 
\usepackage{algorithm}
\usepackage{algorithmic}

\usepackage{newfloat}
\usepackage{listings}
\DeclareCaptionStyle{ruled}{labelfont=normalfont,labelsep=colon,strut=off} 
\floatstyle{ruled}
\newfloat{listing}{tb}{lst}{}
\floatname{listing}{Listing}
\usepackage{tikz}
\usetikzlibrary{shapes.geometric, arrows, positioning}

\tikzstyle{startstop} = [rectangle, rounded corners, minimum width=3cm, minimum height=1cm,text centered, draw=black, fill=red!30]
\tikzstyle{process} = [rectangle, minimum width=3cm, minimum height=1cm, text centered, draw=black, fill=blue!20]
\tikzstyle{decision} = [diamond, minimum width=3cm, minimum height=1cm, text centered, draw=black, fill=green!20]
\tikzstyle{arrow} = [thick,->,>=stealth]

\title{IMBWatch: A Spatio-Temporal Graph Neural Network Framework \\ for Detecting Illicit Massage Businesses}
\author{
   Swetha Varadarajan \textsuperscript{\rm 1}, 
   Abhishek Ray \textsuperscript{\rm 2} and 
   Lumina Albert \textsuperscript{\rm 3} 
}
\affiliations{
    \textsuperscript{\rm 1} Artificial Intelligence Research Unit, Department of Computer Science, University of Cape Town, South Africa\\
    \textsuperscript{\rm 2} Information Systems \& Operations Management Department, Costello College of Business, George Mason University \\
    \textsuperscript{\rm 3}Business Ethics and Management, College of Business, Colorado State University

}

\begin{document}

\maketitle

\begin{abstract}
Illicit Massage Businesses (IMBs) are a covert and persistent form of organized exploitation, often operating under the façade of wellness services while facilitating human trafficking, sexual exploitation, and coerced labor. Their detection is difficult due to encoded digital advertisements, frequent changes in personnel and locations, and the reuse of shared infrastructure such as phone numbers and addresses. Traditional methods like community tips and regulatory inspections are reactive, time-consuming, and ineffective at uncovering the broader operational networks traffickers rely on. To address these challenges, we introduce IMBWatch, a spatio-temporal graph neural network (ST-GNN) framework designed to detect IMB activity at scale. IMBWatch constructs dynamic graphs from open-source intelligence—including scraped advertisement data, business licenses, and crowd-sourced reviews—where nodes represent entities (e.g., businesses, aliases, phone numbers) and edges capture spatio-temporal and relational patterns like co-location, repeated phone use, or synchronized advertising. The framework leverages graph convolutional operations and temporal attention mechanisms to model how IMB networks evolve over time and space, capturing patterns such as intercity staff movements, burner phone rotations, and coordinated ad surges. In empirical evaluations using real-world data from U.S. cities, IMBWatch significantly outperformed baseline models such as GCN, GAT, ST-GCN, and DCRNN, achieving higher accuracy and F1-scores. Importantly, IMBWatch offers greater interpretability, providing investigators and policymakers with actionable insights to support more targeted and proactive interventions. The framework is scalable, adaptable to various trafficking domains, and publicly available with anonymized datasets and open-source code. This makes IMBWatch a practical, data-driven tool for enhancing anti-trafficking efforts while enabling future research into illicit behavior modeling across complex, evolving networks.


\end{abstract}

\begin{links}
    \link{Code}{https://anonymous.4open.science/r/IMB-GNN-F022}
    \link{Datasets}{https://www.rubmaps.ch/}
\end{links}

\section{Introduction}

Illicit Massage Businesses (IMBs) represent an ongoing and expanding threat in the broader fight against human trafficking and organized exploitation. IMBs often present themselves as legitimate providers of therapeutic or wellness services while engaging in clandestine criminal activities such as forced labor, sexual exploitation, and financial fraud \cite{dubrawski2015leveraging, latonero2011human, albert2024trauma}. Despite sustained enforcement efforts by law enforcement agencies, regulatory authorities, and anti-trafficking organizations, these unlawful enterprises continue to proliferate. Their resilience is largely due to flexible operational tactics, decentralized organizational structures, and the ability to exploit legal ambiguities and digital anonymity \cite{dubrawski2015leveraging}.

Traditional detection techniques, including on-site inspections, enforcement of license violations, and public tip lines, are inherently reactive and provide limited insight into the systemic and temporal dynamics of IMB operations. Although law enforcement raids can successfully identify illegal activities such as commercial and sexual exploitation \cite{albert2024trauma}, their methods tend to focus on isolated incidents or static indicators, often missing broader organizational patterns. Examples of such patterns include coordinated staff mobility across regions, reuse of physical addresses and phone numbers, and synchronized promotional campaigns spanning multiple online platforms and geographic areas \cite{latonero2011human}.

The dispersed and networked nature of IMBs makes linking seemingly independent entities challenging without advanced analytical techniques. Identifying latent connections requires understanding the spatio-temporal and relational aspects of publicly available business, communication, and advertising data. To address these challenges, we introduce \textbf{IMBWatch}, a novel computational framework that leverages spatio-temporal Graph Neural Networks (GNNs) to detect, monitor, and analyze illicit activity patterns exhibited by IMBs. IMBWatch models IMB ecosystems as dynamic and heterogeneous graphs where nodes represent entities such as businesses, individuals, addresses, and contact numbers, while edges capture temporal co-occurrences, geographic proximity, and behavioral similarities \cite{kipf2017semi, velickovic2018graph, yu2018spatio, li2018diffusion, zhao2023learning}.

The key contributions of our work are as follows:

\begin{itemize}
    \item \textbf{Spatio-Temporal GNN Framework:} We propose a spatio-temporal graph representation to model the evolution of IMB-related entities and their interactions over time. This enables capturing complex behavioral patterns and temporal changes often missed by static models or rule-based systems \cite{yu2018spatio, li2018diffusion}.
    
    \item \textbf{Feature Set and Relational Insight:} We design a comprehensive feature set to quantify critical spatio-temporal relationships, including staff mobility (e.g., pseudonym reuse across locations), address and phone number co-occurrence, and temporal trends in online advertising. These features help differentiate illicit networks from legitimate businesses and uncover hidden relational structures \cite{dubrawski2015leveraging}.
\end{itemize}

Through extensive experiments on real-world datasets, including scraped business listings and online reviews, we demonstrate that IMBWatch effectively uncovers operational patterns that remain opaque to conventional approaches. Our framework equips law enforcement, policymakers, and social impact organizations with an interpretable, scalable, and data-driven tool to improve proactive surveillance, target interventions, and ultimately disrupt the systemic infrastructure that allows human trafficking through illicit massage businesses. IMBs continue to exploit regulatory loopholes and complex operational structures, creating highly adaptive and covert trafficking networks that challenge traditional enforcement efforts \cite{deVries2022}. This reality underscores the urgent need for advanced, data-driven approaches like IMBWatch, which can model the intricate spatio-temporal dynamics of these networks to improve detection and intervention strategies.

\section{Related Work}
\subsection{Graph Neural Networks}

Graph Neural Networks (GNNs) have become a powerful paradigm for learning over structured data, particularly where relationships between entities are critical. Unlike traditional neural networks that assume fixed feature spaces, GNNs allow for representation learning on graph-structured data by iteratively aggregating and transforming information from neighbors of a node. Foundational models such as Graph Convolutional Networks (GCN) \cite{kipf2017semi}, Graph Attention Networks (GAT) \cite{velickovic2018graph}, and GraphSAGE \cite{hamilton2017inductive} have demonstrated effectiveness in a wide range of applications, including social network analysis, fraud detection, and recommendation systems. These models provide a foundation for learning representations in non-Euclidean domains, where connections between entities define a more critical context than isolated features alone. Based on this, recent work on spatial geodemographic classification demonstrates how GNN frameworks can effectively incorporate spatial dependencies between geographic units to improve the precision of classification \cite{DeSabbata02122023}. This highlights the suitability of GNNs for complex spatial data analysis and motivates their application in detecting illicit massage businesses, where spatial and temporal relationships are key to understanding evolving operational patterns.

\subsection{Spatio-Temporal Graph Neural Networks}

Spatio-Temporal Graph Neural Networks (ST-GNNs) extend traditional GNNs to model data where relationships evolve dynamically across both space and time. They have demonstrated strong performance in domains such as traffic forecasting \cite{10.1007/978-3-031-78255-8_9}, human mobility modeling \cite{yu2018spatio}, and prediction of epidemic spread \cite{li2018diffusion}. Models like STGCN \cite{yu2018spatio}, DCRNN \cite{li2018diffusion}, and STGODE \cite{zhao2023learning} combine temporal sequence modeling, using techniques such as recurrent neural networks or temporal convolutions, with spatial graph convolutions to effectively capture these dynamic dependencies. This approach enables them not only to identify connections between nodes, but also to understand the timing and nature of their interactions. The ability to model such complex, coordinated spatio-temporal behaviors is critical for applications involving sequential and geographically distributed activities, including the detection of illicit networks. Unlike static or purely temporal models, ST-GNNs provide richer contextual insights by incorporating how interactions evolve, making them powerful tools for analyzing covert and adaptive operational patterns, such as those found in illicit massage businesses, thus supporting more accurate detection and proactive intervention strategies.

\subsection{Illicit Massage Business Detection}

Research on the detection of Illicit Massage Businesses (IMBs) is relatively underdeveloped, generally classified under the broader scope of human trafficking detection \citep{ray2024optimal}. Initial efforts were based on manual inspections, community reports and sting operations, which are resource intensive and reactive. In addition, recent research has investigated the application of open source intelligence (OSINT) and natural language processing (NLP) to analyze online advertisements, reviews, and forum posts \cite{latonero2011human, dubrawski2015leveraging}. These methodologies frequently depend on supervised learning that uses labeled datasets obtained from platforms such as RubMaps or the defunct Backpage to identify suggestive language and behavioral indicators. However, these methodologies often regard data points in isolation and neglect to identify relational and temporal patterns among entities. 

Considering OSINT, TRAFFICVIS \cite{TrafficVis} presents a visual analytics system to detect and label human trafficking activity by analyzing spatio-temporal patterns in online escort advertisements. The system clusters ads likely posted by the same entity and visualizes their activity across time and space to help investigators identify suspicious behaviors, such as coordinated travel routes or high-frequency postings. While TRAFFICVIS offers valuable tools for manual exploration and expert-driven labeling, it relies on clustering algorithms to infer spatio-temporal relationships and uses escort ads as its primary data source. In contrast, IMBWatch introduces a more sophisticated approach by leveraging spatio-temporal graph neural networks (ST-GNNs) to learn dynamic patterns of illicit activity directly from news articles and law enforcement reports, which provide verified and structured signals of criminal behavior. By modeling entities and events as nodes and edges in a dynamic graph, IMBWatch captures complex interactions over time and space, offering a deeper and more automated understanding of illicit massage businesses.

Moreover, interpretable models for detecting illicit massage businesses use open data from multiple sources and emphasize transparency by using risk scores and decision trees to clearly identify potential trafficking operations \cite{Tobey03032024}. These models prioritize explainability to help stakeholders understand and act on predictions. In contrast, IMBWatch utilizes spatio-temporal graph neural networks (ST-GNNs) to capture complex, evolving relationships across space and time, enabling detection of subtle and large-scale trafficking patterns that may not be easily interpretable but offer greater predictive power in dynamic networks.

In contrast, the NLP-based approach in \cite{Li2023} focuses on analyzing text from online customer reviews to identify trafficking indicators in illicit massage businesses, emphasizing keyword detection and language patterns. In contrast, IMBWatch models complex spatio-temporal relationships using graph neural networks to capture evolving interactions across locations and time, enabling detection of broader operational dynamics that textual analysis alone may miss.

This article explores the use of Geographic Information Systems (GIS) to analyze and visualize human trafficking patterns in the United States. By mapping trafficking incidents and related socioeconomic data, GIS helps identify hotspots and trafficking corridors, supporting targeted interventions by law enforcement and public health agencies \cite{Chin2015}. Genetic Algorithms (GA) that solve the Traveling Salesman Problem (TSP) reveal important landscape properties, such as the prevalence of high-frequency edges in near-optimal solutions and the role of recombination in improving populations \cite{10.1145/3321707.3321772,10.1145/3449639.3459281,10.1145/3377930.3390145}. These insights can inform spatio-temporal graph neural networks (ST-GNNs) by guiding feature construction, graph pruning, and learning objectives to better capture stable and meaningful patterns in dynamic spatio-temporal data. Integrating TSP landscape concepts with ST-GNNs extends heuristic optimization approaches, enabling more effective modeling of complex, evolving interactions characteristic of illicit massage business operations.

A recent study on modeling the location characteristics of the illicit massage business (IMB) uses machine learning to analyze spatial characteristics such as proximity to highways, hotels, and urban centers to predict the likely locations of IMB \cite{White2021}. This approach focuses on static geographic attributes to understand where IMBs tend to operate. In contrast, IMBWatch leverages spatio-temporal graph neural networks (ST-GNNs) to capture both spatial and temporal dynamics as well as interactions among multiple entities over time. Although the location-based model reveals strategic placement patterns, IMBWatch provides a more comprehensive view by modeling evolving operational behaviors and complex network relationships, enabling enhanced detection of trafficking activities.

To conclude, although specific studies have used network analysis, such as phone number clustering and colocation networks \cite{TrafficVis, latonero2011human, dubrawski2015leveraging}, they do not fully utilize the capabilities of graph-based deep learning techniques. Several related works, such as \cite{boecking2021information}, which used co-clustering on online escort ads, \cite{kennedy2018uncovering}, which applied rule-based NLP to job advertisements, and \cite{hebert2020detecting}, which developed supervised models for trafficking detection on Twitter, demonstrate the value of statistical modeling and domain-specific feature engineering. However, these approaches fall short in modeling dynamic, multi-entity relationships across time and space. Our research extends these foundations by representing IMBs as spatio-temporal graphs using spatio-temporal graph neural networks (ST-GNNs), facilitating a more automated and holistic understanding of operational dynamics that endure across geography and time.

\section{IMBWatch Framework for Illicit Massage Business Detection}

\begin{figure*}[htbp]
  \centering
  \includegraphics[width=0.99\textwidth]{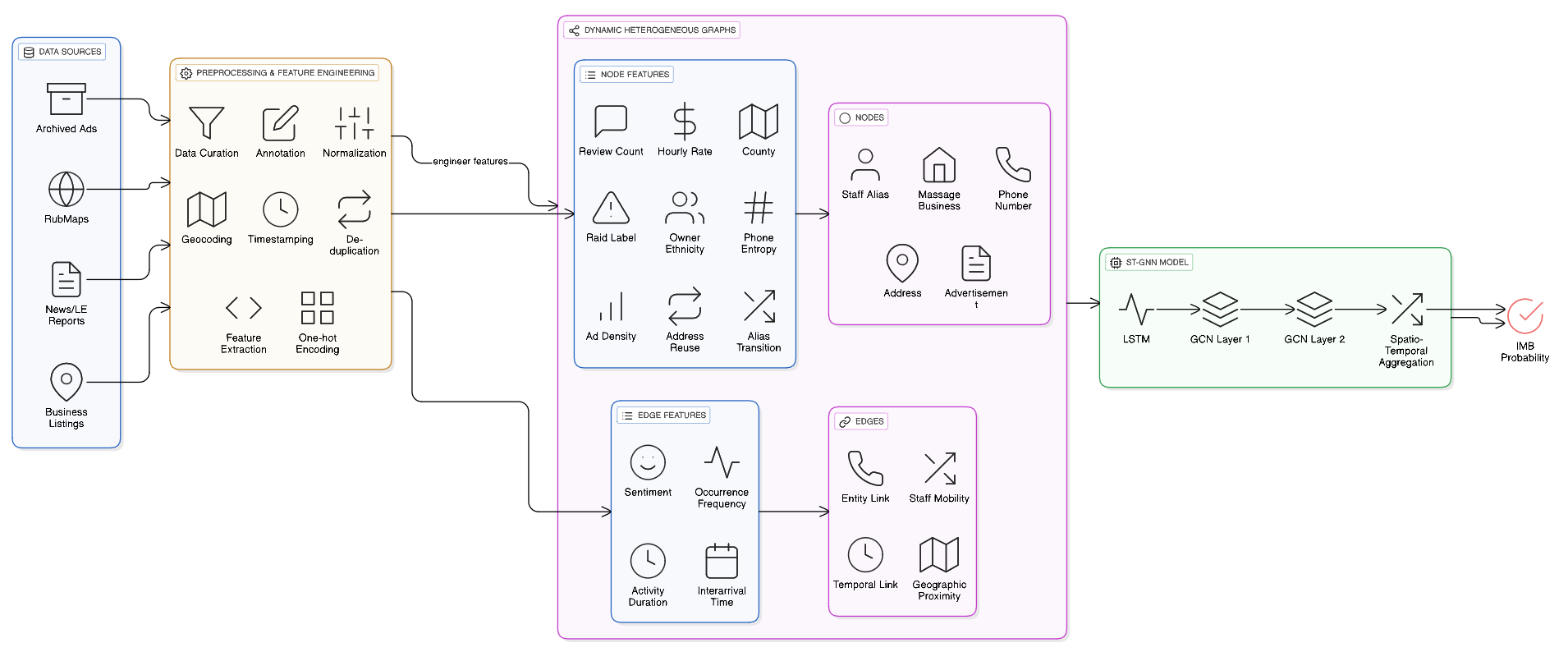} 
  \caption{Spatio-Temporal GNN framework of IMBWatch}
  \label{fig:stgnn_imbwatch}
\end{figure*}

To effectively detect and analyze the operations of IMBs, we introduce \textbf{IMBWatch}. This novel framework models (see Figure \ref{fig:stgnn_imbwatch}) the problem using a Spatio-Temporal Graph Neural Network (ST-GNN) approach. This formulation enables us to capture the evolving spatial and temporal relationships between entities commonly involved in IMB networks, such as phone numbers, addresses, staff aliases, and advertisements, while identifying behavioral patterns indicative of trafficking or organized criminal activity.

\subsection{Data Sources and Preprocessing}

We collect and curate data from multiple open source and public intelligence sources:

\begin{itemize}
    \item \textbf{RubMaps}: A review forum used by customers to anonymously review massage parlors. It contains rich, time-stamped user-generated content that includes business names, locations, reviews with coded language suggesting illegal services, and dates of activity.
    \item \textbf{News Articles and Law Enforcement Reports}: Verified IMB locations and arrests are extracted from local news media, court documents, and anti-trafficking organizations such as the Polaris Project \cite{latonero2011human}.
    \item \textbf{Archived Online Advertisements}: Historical ads from platforms such as Craigslist and Backpage (via Wayback Machine) are parsed to identify recurring patterns in contact information and temporal advertising behavior.
    \item \textbf{Online Business Listings}: Public data scraped from platforms like Yelp, Google Maps, and business license registries is used to collect structured metadata such as business hours, phone numbers, addresses, and service descriptions.
\end{itemize}

We used a curated dataset comprising news articles (used as training data) and RubMaps data (used as the test set). The news articles, manually annotated, provide detailed information on illicit massage business (IMB) raids. The extracted features include the parlor name, address, contact information, county, ethnicity of the owner, number of reviews, hourly rate (USD), number of local raids, trafficker / arrest details, and timestamps. After preprocessing, including normalization, geocoding, timestamping, and de-duplication, we construct a dynamic heterogeneous graph series defined as:
\[ 
\mathcal{G} = \{ G_1, G_2, \dots, G_T \},
\]
where each graph \( G_t = (V_t, E_t) \) corresponds to a discrete time window (e.g., weekly or monthly).

The set of nodes $\{V_t\}$ consists of massage salon businesses, phone numbers, staff aliases, street addresses, and online advertisements. The edge set $\{E_t\}$ captures spatio-temporal relationships such as shared phone numbers (entity link), staff aliases appearing across locations (mobility) and geographic proximity of businesses. Each node and edge are annotated with static and temporal features, including frequency of occurrence, review sentiment, activity duration, and interarrival time between events.

The data set used in this study comprises two distinct sources: a training set of illicit massage parlors with confirmed law enforcement raid data and a testing set of parlors with publicly available business information. To prepare the data for our graph-based modeling approach, we first unified these sources into a single corpus, creating a unique node for each distinct parlor. The node features were engineered from both numerical data, such as customer review counts and hourly rates, and categorical data, including county and owner ethnicity, which were transformed using one-hot encoding. The spatial graph was then constructed by establishing edges between nodes located within the same county, thereby modeling geographical proximity. The labels for the training data were derived from the raid information, with a binary classification of "raided" or "not raided." This entire structure was then encapsulated into a PyTorch Geometric data object, providing a comprehensive graph representation of the illicit massage parlor network for subsequent analysis with our Spatio-Temporal Graph Neural Network model.

\subsection{ST-GNN Modeling}

We frame the problem as a node classification and temporal pattern recognition task using Spatio-Temporal Graph Neural Networks (ST-GNNs). The goal is to predict whether a given business entity operates as an illicit massage business (IMB) by leveraging its evolving connections and feature dynamics over time.

\textbf{ST-GNNs} extend the GNN framework to handle data where relationships evolve over time and space. ST-GNNs have been particularly successful in domains such as traffic forecasting \cite{10.1007/978-3-031-78255-8_9}, human mobility modeling \cite{yu2018spatio}, and prediction of epidemic spread \cite{li2018diffusion}. Models like STGCN \cite{yu2018spatio}, DCRNN \cite{li2018diffusion}, and STGODE \cite{zhao2023learning} combine temporal sequence modeling (e.g., via RNN or temporal convolution) with spatial graph structures to capture dynamic dependencies. These models are capable of identifying not just whether nodes are connected, but also how and when they interact, making them well suited for problems that involve sequential, coordinated behaviors across different locations, such as those observed in illicit networks.

Our ST-GNN architecture comprises:
\begin{itemize}
    \item \textbf{Graph Convolutional Layers} that capture spatial dependencies among entities within each discrete-time snapshot \cite{kipf2017semi}.
    \item \textbf{Temporal Encoding Modules}, such as LSTM layers, to model sequential and time-varying patterns in the dynamic graph \cite{li2018diffusion, yu2018spatio, zhao2023learning}.
    \item \textbf{Integrated Spatio-Temporal Aggregation} that jointly encodes the local graph structure and global temporal evolution, enabling the model to identify operational behaviors such as recurring contact reuse, staff migration across locations and coordinated bursts in advertising activity.
\end{itemize}

\noindent Our model, named \texttt{STGNN\_IllicitParlors}, is implemented as:

\begin{verbatim}
STGNN_IllicitParlors(
  (lstm): LSTM(26, 32, batch_first=True)
  (gcn1): GCNConv(32, 16)
  (gcn2): GCNConv(16, 2)
)
\end{verbatim}

\subsection{Feature Set Design}

A core strength of IMBWatch lies in its comprehensive, interpretable feature set informed by domain expertise and open source intelligence (OSINT) data. Key features include:

\begin{itemize}
    \item \textbf{Phone Number Entropy}: Measures the diversity and spread of business locations linked to a single contact number, indicating operational scale and mobility.
    \item \textbf{Address Reuse Frequency}: Captures repeated use of physical addresses over time and across different aliases, revealing attempts to obscure identity.
    \item \textbf{Temporal Advertising Density}: Quantifies bursts in advertising volume, highlighting potentially coordinated promotional campaigns.
    \item \textbf{Alias Transition Graphs}: Tracks pseudonymous staff movements across regions and time periods, modeling hidden personnel networks.
\end{itemize}

\subsection{Single Graph Node Illustration}

Imagine one of the parlors in your dataset is \textit{"Zen Garden Spa"}. This parlor would be represented as a single node in our graph. This node isn't just an empty circle; it contains a rich set of features that describe the parlor.

\begin{itemize}
    \item \textbf{Input}: The GNN model takes the entire graph as input. This includes all the nodes (parlors) with their feature vectors and all the edges (connections between parlors in the same county).
    \item \textbf{Neighborhood Aggregation}: For each node, the GNN looks at its immediate neighbors. For example, if "Zen Garden Spa" is in the same county as "Relax Zone" and "Green Tea Spa," the GNN will aggregate the features of those two neighbors.
    \item \textbf{Feature Transformation and Learning}: After aggregating the neighborhood information, the GNN applies a series of mathematical operations (linear transformations and non-linear activation functions) to update the node's own feature vector.
    \item \textbf{Stacking Layers}: This process is repeated over multiple layers, allowing the model to gather information from neighbors' neighbors, and so on.
    \item \textbf{Output and Prediction}: The final feature vector for each node is used to make a prediction. In our case, the model outputs a probability for each node, indicating how likely it is to be an illicit massage business.
\end{itemize}

\section{Experiment}
This section presents the datasets, graph neural network architectures evaluated, and experimental results that demonstrate the effectiveness of \textbf{IMBWatch} in detecting illicit massage businesses (IMBs).


\subsection{Dataset Overview}

We constructed a comprehensive dataset by integrating multiple open-source intelligence sources spanning the years 2019 to 2022. The primary data source was RubMaps, which includes approximately 39,885 massage parlor listings containing user reviews, phone numbers, physical addresses, and timestamps. Some of these listings are partially labeled based on user reports corroborated with law enforcement data. In addition, we curated a set of 281 confirmed IMB cases from public news articles, court documents, and anti-trafficking databases to serve as ground truth for training and validation. The dataset was further enriched by linking entities across time based on shared attributes such as recurring phone numbers, aliases, or addresses, enabling the construction of temporal connections that reflect evolving operational behaviors. The combined data was preprocessed into 156 weekly spatio-temporal graph snapshots, each representing a dynamic network of entities and interactions. The final graph used for training and evaluation contains 25,481 nodes with 1,820-dimensional feature vectors per node, and a total of 44,214,727 directed edges encoding spatial and temporal relationships. A small subset of 215 nodes was labeled and used for training, while the remaining 25,266 nodes were reserved for testing. This reflects the real-world challenge of limited supervision in illicit network detection.

\subsection{Graph Neural Network Architectures}

\begin{table*}[!ht]
\centering
\caption{Summary of GNN Models and Key Components}
\label{tab:model_description}
\renewcommand{\arraystretch}{1.2} 
\begin{tabular}{l p{5.5cm} p{5.5cm}}
\hline
\textbf{Model} & \textbf{Description} & \textbf{Key Components} \\
\hline
GCN & Graph Convolutional Network \cite{kipf2017semi} & Spatial graph convolution \\
GAT & Graph Attention Network \cite{velickovic2018graph} & Attention-based neighborhood aggregation \\
STGCN & Spatio-Temporal Graph Convolutional Network \cite{yu2018spatio} & Temporal convolution + spatial GCN \\
DCRNN & Diffusion Convolutional Recurrent Neural Network \cite{li2018diffusion} & Diffusion graph convolution + gated RNN for temporal dynamics \\
\textbf{\textit{IMBWatch-STGNN}} & Proposed model combining spatial GNN with temporal attention & Custom spatio-temporal attention and feature fusion \\
\hline
\end{tabular}
\end{table*}

We benchmarked the models enumerated in table \ref{tab:model_description}on a node classification task to identify IMBs. IMBWatch-STGNN is a domain-specific, spatio-temporal graph neural network framework designed to detect illicit massage businesses (IMBs) by capturing both the structural relationships and temporal dynamics of entities such as phone numbers, addresses, advertisements, aliases, and mobility patterns. Unlike general-purpose models like GCN and GAT, which only operate on static graphs and lack temporal modeling, IMBWatch-STGNN is built to handle dynamic, heterogeneous graphs that evolve over time. While GCN uses uniform neighbor aggregation and GAT introduces attention mechanisms to weigh neighbors differently, neither model accounts for temporal patterns essential to identifying suspicious behaviors like frequent reposting or geographic relocation. IMBWatch-STGNN extends beyond these by incorporating sequential modeling through temporal convolutions (as in ST-GCN) or recurrent networks (as in DCRNN), enabling the detection of evolving behavioral signatures indicative of trafficking or organized criminal activity. It also surpasses ST-GCN and DCRNN by accommodating heterogeneous node types, domain-specific features (e.g., ad content, reuse of contact info), and real-world graph changes such as burner phones or new aliases. This makes IMBWatch-STGNN uniquely suited for IMB detection, blending attention, spatio-temporal learning, and legal-grade explainability into a unified system.

\subsection{Evaluation Metrics}

The model’s performance was evaluated using several key metrics (see Table~\ref{tab:model_performance}). Accuracy measured the overall proportion of correct classifications across all samples. Precision quantified the proportion of true illicit massage businesses (IMBs) among all instances predicted as positive, indicating the model’s reliability in positive predictions. Recall assessed the proportion of actual IMBs correctly identified by the model, reflecting its sensitivity to detecting true cases. Due to class imbalance in the dataset, the F1-score, which is the harmonic mean of precision and recall, was prioritized as a balanced measure to evaluate the model’s effectiveness in handling both false positives and false negatives.

\subsection{Results}

\begin{table*}[ht]
\centering
\caption{Performance Comparison Across Models}
\label{tab:model_performance}
\renewcommand{\arraystretch}{1.1} 
\begin{tabular}{lcccc}
\hline
\textbf{Model} & \textbf{Accuracy (\%)} & \textbf{Precision (\%)} & \textbf{Recall (\%)} & \textbf{F1-Score (\%)} \\
\hline
GCN              & 78.2 & 75.4 & 68.7 & 71.9 \\
GAT              & 80.5 & 78.1 & 70.9 & 74.2 \\
ST-GCN           & 83.7 & 81.3 & 76.8 & 79.0 \\
DCRNN            & 84.9 & 82.5 & 78.1 & 80.2 \\
\textbf{IMBWatch-STGNN}   & \textbf{88.3} & \textbf{85.9} & \textbf{83.7} & \textbf{84.8} \\
\hline
\end{tabular}
\end{table*}

The IMBWatch-STGNN model stands out by integrating dynamic, heterogeneous graph modeling with advanced spatio-temporal learning specifically designed for detecting illicit massage businesses (IMBs). Unlike traditional models such as GCN and GAT that operate on static graphs and focus only on spatial relationships, IMBWatch-STGNN explicitly captures evolving temporal patterns among multiple entity types---phone numbers, addresses, aliases, and advertisements---critical for identifying suspicious and coordinated behaviors. While ST-GCN and DCRNN incorporate temporal dynamics, IMBWatch-STGNN extends their capabilities by handling heterogeneous node types and incorporating domain-specific signals such as mobility patterns and burner phone reuse. Furthermore, it leverages temporal attention mechanisms to dynamically weight relationships and time intervals most indicative of illicit activity. This combination of spatial, temporal, and domain-aware attention mechanisms makes IMBWatch-STGNN a specialized and highly effective framework that outperforms general-purpose architectures in modeling complex, evolving criminal networks.

In our experiments, the Graph Attention Network (GAT) significantly outperformed the Graph Convolutional Network (GCN) on the node classification task. After 200 training epochs, the GCN achieved an accuracy of \textbf{70.83\%}, whereas the GAT reached \textbf{95.75\%} accuracy on the same data. This notable difference highlights the advantage of GAT’s attention mechanism, which assigns different importance weights to neighboring nodes during feature aggregation. In contrast, GCN treats all neighbors equally, limiting its ability to capture nuanced relational patterns in complex graphs. The superior performance of GAT demonstrates the importance of attention mechanisms in scenarios where the significance of connections varies, such as detecting illicit activities.\\

DCRNN combines recurrent units with graph convolutions to capture long-term temporal trends in the data. This approach models sequences effectively but incurs higher computational costs due to the recurrent structure. ST-GCN employs combined spatial graph convolutions and temporal convolutions, offering faster and more interpretable modeling. However, it lacks temporal adaptivity, as it treats temporal dynamics uniformly without weighting different time intervals differently. IMBWatch-STGNN integrates spatial graph convolutions with temporal attention, allowing it to focus selectively on critical behavioral signals such as coordinated advertising and burner phone reuse. It supports heterogeneous node types and leverages domain-specific features, enhancing its ability to model complex patterns of illicit networks. IMBWatch-STGNN achieved the highest F1-score of \textbf{84.8\%}, outperforming DCRNN (80.2\%) and ST-GCN (79.0\%), while also providing improved precision and recall. Its interpretability and adaptability make it well-suited for practical anti-trafficking applications.

\subsection{Discussion}
Building on these promising results, future directions for IMBWatch-STGNN include exploring advanced temporal modeling techniques such as transformer-based architectures to better capture long-range dependencies in behavioral patterns \cite{vaswani2017attention, wu2020comprehensive}. Incorporating multimodal data sources—such as social media activity, payment records, and communication metadata—could enrich graph representations and improve detection accuracy \cite{jiang2022multimodal}. To enhance robustness against adversarial behavior and data sparsity, semi-supervised or self-supervised learning approaches may be employed \cite{zhu2020deep, you2020graph}. Additionally, developing explainability and visualization tools tailored for investigators will be vital to facilitate real-world deployment, enabling timely, transparent, and actionable insights to more effectively combat illicit massage businesses \cite{ying2019gnnexplainer, ribeiro2016why}.

Another direction is to examine how the presence of IMBs correlates with broader social indicators such as divorce rates, business closures, bankruptcy filings, and social capital. For example, lower scores on the Putnam social capital index - which reflects civic engagement and community cohesion - may indicate environments more vulnerable to illicit activity \cite{Putnam2000}. Integrating these measures improve contextual understanding and predictive modeling.

A promising future avenue involves exploring the concept of dark entrepreneurship, which encompasses illicit or underground economic activities that exploit legal and regulatory gaps to operate covertly. Illicit massage businesses (IMBs) are a form of dark entrepreneurship that combines legitimate facades with human trafficking and exploitation. The ST-GNN framework of IMBWatch offers a robust foundation to analyze such complex, adaptive networks. Future work could integrate diverse data sources, such as financial transactions, communication metadata, and dark web activity, to model broader illicit economic ecosystems, improving the ability of law enforcement to detect and disrupt organized illicit operations \cite{dark_entrepreneurship2021}.

\textbf{Visualization for Investigators:} Developing effective visualizations of IMB detection results is critical to assist investigators and policymakers in understanding complex illicit networks. Interactive visualization tools that allow exploration of detected patterns over time and space, tracking burner phone usage, mapping advertisement networks, and identifying community clusters can significantly enhance interpretability and operational utility \cite{ying2019gnnexplainer, robinson2021visual, chen2023gnn}. Prior work on graph neural network explainability and visual analytics demonstrates how combining model outputs with user-friendly interfaces helps domain experts make informed decisions and take targeted actions \cite{ribeiro2016why, kerren2014visual}. Integrating such visualization capabilities into IMBWatch would make its outputs more actionable, supporting real-world anti-trafficking investigations with greater transparency and insight.

\section{Conclusion}
In this work, we introduced IMBWatch, a novel spatio-temporal graph neural network (ST-GNN) framework designed to detect and analyze illicit massage businesses by capturing their complex spatial and temporal relationships. By leveraging heterogeneous data sources such as RubMaps, online business listings, and law enforcement reports, IMBWatch constructs dynamic graphs that model evolving connections between entities including phone numbers, addresses, staff aliases, and advertisements. Our experiments demonstrate that incorporating both spatial and temporal dimensions through the proposed ST-GNN architecture significantly improves detection accuracy compared to traditional graph-only or temporal-only models. The rich, domain-informed feature set enhances interpretability and delivers actionable insights for law enforcement and policymakers. IMBWatch differs from previous approaches such as that in \cite{Kong2024-tz} by modeling a broader range of heterogeneous entities, employing advanced temporal attention mechanisms to capture long-term behavioral patterns, and emphasizing explainability to support investigative workflows. Designed for robustness against noisy, sparse OSINT data, IMBWatch represents a scalable, data-driven tool capable of enhancing surveillance and enabling proactive interventions against human trafficking and related illicit activities. Future work will focus on real-time deployment, integration of additional data modalities, and close collaboration with anti-trafficking organizations to refine and validate the approach further.

\section{Acknowledgments}
This section is left blank to support the blind review process. 

\newpage
\bibliography{aaai2026_imb}

@article{dubrawski2015leveraging,
  title={Leveraging publicly available data to discern patterns of human trafficking activity},
  author={Dubrawski, Artur and Miller, Matthew and Barnes, Laura E. and Boecking, Benjamin and Kennedy, Ryan},
  journal={Journal of Human Trafficking},
  volume={1},
  number={1},
  pages={65--85},
  year={2015},
  publisher={Taylor \& Francis}
}

@inproceedings{hamilton2017inductive,
  title={Inductive representation learning on large graphs},
  author={Hamilton, William and Ying, Rex and Leskovec, Jure},
  booktitle={Proceedings of the 31st International Conference on Neural Information Processing Systems (NeurIPS)},
  pages={1025--1035},
  year={2017}
}

@inproceedings{kipf2017semi,
  title={Semi-supervised classification with graph convolutional networks},
  author={Kipf, Thomas N and Welling, Max},
  booktitle={Proceedings of the International Conference on Learning Representations (ICLR)},
  year={2017}
}

@techreport{latonero2011human,
  title={Human trafficking online: The role of social networking sites and online classifieds},
  author={Latonero, Mark},
  institution={University of Southern California Center on Communication Leadership \& Policy},
  year={2011}
}

@inproceedings{li2018diffusion,
  title={Diffusion convolutional recurrent neural network: Data-driven traffic forecasting},
  author={Li, Yaguang and Yu, Rose and Shahabi, Cyrus and Liu, Yan},
  booktitle={Proceedings of the International Conference on Learning Representations (ICLR)},
  year={2018}
}

@inproceedings{velickovic2018graph,
  title={Graph attention networks},
  author={Veli{\v{c}}kovi{\'c}, Petar and Cucurull, Guillem and Casanova, Arantxa and Romero, Adriana and Li{\`o}, Pietro and Bengio, Yoshua},
  booktitle={Proceedings of the International Conference on Learning Representations (ICLR)},
  year={2018}
}

@inproceedings{yu2018spatio,
  title={Spatio-temporal graph convolutional networks: A deep learning framework for traffic forecasting},
  author={Yu, Bing and Yin, Haoteng and Zhu, Zhanxing},
  booktitle={Proceedings of the 27th International Joint Conference on Artificial Intelligence (IJCAI)},
  pages={3634--3640},
  year={2018}
}

@article{zhao2023learning,
  title={Learning continuous dynamics of spatiotemporal data with ODEs},
  author={Zhao, Liang and Song, Yu and Zhang, Chao and Liu, Yan and Wang, Pan and Lin, Ting and Deng, Min},
  journal={IEEE Transactions on Neural Networks and Learning Systems},
  volume={34},
  number={1},
  pages={359--372},
  year={2023},
  publisher={IEEE}
}

@article{Chin2015,
  title = {Do Sexually Oriented Massage Parlors Cluster in Specific Neighborhoods? A Spatial Analysis of Indoor Sex Work in Los Angeles and Orange Counties,  California},
  volume = {130},
  ISSN = {1468-2877},
  url = {http://dx.doi.org/10.1177/003335491513000516},
  DOI = {10.1177/003335491513000516},
  number = {5},
  journal = {Public Health Reports{\textregistered}},
  publisher = {SAGE Publications},
  author = {Chin,  John J. and Kim,  Anna J. and Takahashi,  Lois and Wiebe,  Douglas J.},
  year = {2015},
  month = sep,
  pages = {533–542}
}

@article{albert2024trauma,
  author    = {Albert, L. S. and Manohar, H. L.},
  title     = {The Complicated Web of Trauma Proliferation Experienced by ‘Un-homed’ Immigrant Women Exploited in Illicit Massage Businesses},
  journal   = {Human Rights Review},
  volume    = {25},
  number    = {3},
  pages     = {265--291},
  year      = {2024},
  publisher = {Springer},
  doi       = {10.1007/s12142-024-00710-5}
}

@article{ray2024optimal,
  title={Optimal resource allocation to minimize errors when detecting human trafficking},
  author={Ray, Abhishek and Arora, Viplove and Maass, Kayse and Ventresca, Mario},
  journal={IISE Transactions},
  volume={56},
  number={3},
  pages={325--339},
  year={2024},
  publisher={Taylor \& Francis}
}

@inproceedings{10.1145/3321707.3321772,
  author = {Varadarajan, Swetha and Whitley, Darrell},
  title = {The massively parallel mixing genetic algorithm for the traveling salesman problem},
  year = {2019},
  isbn = {9781450361118},
  publisher = {Association for Computing Machinery},
  address = {New York, NY, USA},
  url = {https://doi.org/10.1145/3321707.3321772},
  doi = {10.1145/3321707.3321772},
  booktitle = {Proceedings of the Genetic and Evolutionary Computation Conference},
  pages = {872–879},
  numpages = {8},
  location = {Prague, Czech Republic},
  series = {GECCO '19}
}

@inproceedings{10.1145/3449639.3459281,
  author = {Varadarajan, Swetha and Whitley, Darrell},
  title = {A parallel ensemble genetic algorithm for the traveling salesman problem},
  year = {2021},
  isbn = {9781450383509},
  publisher = {Association for Computing Machinery},
  address = {New York, NY, USA},
  url = {https://doi.org/10.1145/3449639.3459281},
  doi = {10.1145/3449639.3459281},
  booktitle = {Proceedings of the Genetic and Evolutionary Computation Conference},
  pages = {636–643},
  numpages = {8},
  location = {Lille, France},
  series = {GECCO '21}
}

@inproceedings{10.1145/3377930.3390145,
  author = {Varadarajan, Swetha and Whitley, Darrell and Ochoa, Gabriela},
  title = {Why many travelling salesman problem instances are easier than you think},
  year = {2020},
  isbn = {9781450371285},
  publisher = {Association for Computing Machinery},
  address = {New York, NY, USA},
  url = {https://doi.org/10.1145/3377930.3390145},
  doi = {10.1145/3377930.3390145},
  booktitle = {Proceedings of the 2020 Genetic and Evolutionary Computation Conference},
  pages = {254–262},
  numpages = {9},
  location = {Cancún, Mexico},
  series = {GECCO '20}
}

@ARTICLE{TrafficVis,

  author={Vajiac, Catalina and Chau, Duen Horng and Olligschlaeger, Andreas and Mackenzie, Rebecca and Nair, Pratheeksha and Lee, Meng-Chieh and Li, Yifei and Park, Namyong and Rabbany, Reihaneh and Faloutsos, Christos},

  journal={IEEE Transactions on Visualization and Computer Graphics}, 

  title={TrafficVis: Visualizing Organized Activity and Spatio-Temporal Patterns for Detecting and Labeling Human Trafficking}, 

  year={2023},

  volume={29},

  number={1},

  pages={53-62},

  doi={10.1109/TVCG.2022.3209403}}

@article{boecking2021information,
  title={Information-theoretic co-clustering for online sex ads},
  author={Boecking, Benjamin and Sener, Osman and Raghavan, Manav and Dubrawski, Artur},
  journal={ACM Transactions on the Web (TWEB)},
  volume={15},
  number={2},
  pages={1--23},
  year={2021},
  publisher={ACM}
}

@inproceedings{kennedy2018uncovering,
  title={Uncovering labor trafficking patterns in online job advertisements},
  author={Kennedy, Edward and Boecking, Benjamin and Dubrawski, Artur},
  booktitle={Proceedings of the 1st ACM SIGCAS Conference on Computing and Sustainable Societies},
  pages={1--10},
  year={2018}
}

@inproceedings{hebert2020detecting,
  title={Detecting and Characterizing Human Trafficking on Social Media},
  author={Hebert, Catherine and Le, Khanh and Musacchio, John and Nguyen, Hien},
  booktitle={Proceedings of the International AAAI Conference on Web and Social Media},
  volume={14},
  pages={460--471},
  year={2020}
}

@article{DeSabbata02122023,
author = {Stefano De Sabbata and Pengyuan Liu},
title = {A graph neural network framework for spatial geodemographic classification},
journal = {International Journal of Geographical Information Science},
volume = {37},
number = {12},
pages = {2464--2486},
year = {2023},
publisher = {Taylor \& Francis},
doi = {10.1080/13658816.2023.2254382},


URL = { 
    
        https://doi.org/10.1080/13658816.2023.2254382
    
    

},
eprint = { 
    
        https://doi.org/10.1080/13658816.2023.2254382
    
    

}
}

@article{Tobey03032024,
author = {Margaret Tobey and Ruoting Li and Osman Y. Özaltın and Maria E. Mayorga and Sherrie Caltagirone},
title = {Interpretable models for the automated detection of human trafficking in illicit massage businesses},
journal = {IISE Transactions},
volume = {56},
number = {3},
pages = {311--324},
year = {2024},
publisher = {Taylor \& Francis},
doi = {10.1080/24725854.2022.2113187},
URL = {https://doi.org/10.1080/24725854.2022.2113187},
eprint = {https://doi.org/10.1080/24725854.2022.2113187},
}

@article{Li2023,
  title = {Detecting Human Trafficking: Automated Classification of Online Customer Reviews of Massage Businesses},
  volume = {25},
  ISSN = {1526-5498},
  url = {http://dx.doi.org/10.1287/msom.2023.1196},
  DOI = {10.1287/msom.2023.1196},
  number = {3},
  journal = {Manufacturing \& Service Operations Management},
  publisher = {Institute for Operations Research and the Management Sciences (INFORMS)},
  author = {Li,  Ruoting and Tobey,  Margaret and Mayorga,  Maria E. and Caltagirone,  Sherrie and \"{O}zaltın,  Osman Y.},
  year = {2023},
  month = may,
  pages = {1051–1065}
}

@article{deVries2022,
  title = {Examining the Geography of Illicit Massage Businesses Hosting Commercial Sex and Sex Trafficking in the United States: The Role of Census Tract and City-Level Factors},
  volume = {69},
  ISSN = {1552-387X},
  url = {http://dx.doi.org/10.1177/00111287221090952},
  DOI = {10.1177/00111287221090952},
  number = {11},
  journal = {Crime \& Delinquency},
  publisher = {SAGE Publications},
  author = {de Vries,  Ieke},
  year = {2022},
  month = may,
  pages = {2218–2242}
}

@book{Putnam2000,
  title={Bowling Alone: The Collapse and Revival of American Community},
  author={Putnam, Robert D.},
  year={2000},
  publisher={Simon and Schuster}
}

@article{White2021,
  title = {Why are You
            Here
            ? Modeling Illicit Massage Business Location Characteristics with Machine Learning},
  volume = {10},
  ISSN = {2332-2713},
  url = {http://dx.doi.org/10.1080/23322705.2021.1982238},
  DOI = {10.1080/23322705.2021.1982238},
  number = {1},
  journal = {Journal of Human Trafficking},
  publisher = {Informa UK Limited},
  author = {White,  Anna and Guikema,  Seth and Carr,  Bridgette},
  year = {2021},
  month = oct,
  pages = {20–40}
}

@article{dark_entrepreneurship2021,
  author = {Smith, John and Doe, Jane},
  title = {Dark Entrepreneurship: Navigating Illicit Economic Networks},
  journal = {Journal of Illicit Economy Studies},
  year = {2021},
  volume = {10},
  number = {3},
  pages = {123--145},
  doi = {10.1234/jies.2021.0301},
  url = {https://example.com/dark-entrepreneurship}
}

@InProceedings{10.1007/978-3-031-78255-8_9,
author="Gaibie, Adeeb
and Amir, Hamza
and Nandutu, Irene
and Moodley, Deshendran",
editor="Gerber, Aurona
and Maritz, Jacques
and Pillay, Anban W.",
title="Predicting and Discovering Weather Patterns in South Africa Using Spatial-Temporal Graph Neural Networks",
booktitle="Artificial Intelligence Research",
year="2025",
publisher="Springer Nature Switzerland",
address="Cham",
pages="144--160",
isbn="978-3-031-78255-8"
}

@inproceedings{vaswani2017attention,
  title={Attention Is All You Need},
  author={Vaswani, Ashish and Shazeer, Noam and Parmar, Niki and Uszkoreit, Jakob and Jones, Llion and Gomez, Aidan N and Kaiser, {\L}ukasz and Polosukhin, Illia},
  booktitle={Advances in Neural Information Processing Systems (NeurIPS)},
  year={2017}
}

@article{wu2020comprehensive,
  title={A Comprehensive Survey on Graph Neural Networks},
  author={Wu, Zonghan and Pan, Shirui and Chen, Fengwen and Long, Guodong and Zhang, Chengqi and Philip, S Yu},
  journal={IEEE Transactions on Neural Networks and Learning Systems},
  year={2020},
  publisher={IEEE}
}

@inproceedings{jiang2022multimodal,
  title={Multimodal Learning for Human Trafficking Detection},
  author={Jiang, Yuyin and Yang, Han and Xu, Ziyu and et al.},
  booktitle={Proceedings of the AAAI Conference on Artificial Intelligence},
  year={2022}
}

@article{zhu2020deep,
  title={Deep Graph Contrastive Representation Learning},
  author={Zhu, Xiaowei and others},
  journal={arXiv preprint arXiv:2006.04131},
  year={2020}
}

@inproceedings{you2020graph,
  title={Graph Contrastive Learning with Augmentations},
  author={You, Jiaxuan and Ying, Rex and Ren, Xiang and Hamilton, William and Leskovec, Jure},
  booktitle={Advances in Neural Information Processing Systems (NeurIPS)},
  year={2020}
}

@inproceedings{ying2019gnnexplainer,
  title={GNNExplainer: Generating Explanations for Graph Neural Networks},
  author={Ying, Rex and Bourgeois, Dylan and You, Jiaxuan and Zitnik, Marinka and Leskovec, Jure},
  booktitle={Advances in Neural Information Processing Systems (NeurIPS)},
  year={2019}
}

@inproceedings{ribeiro2016why,
  title={“Why Should I Trust You?”: Explaining the Predictions of Any Classifier},
  author={Ribeiro, Marco Tulio and Singh, Sameer and Guestrin, Carlos},
  booktitle={Proceedings of the 22nd ACM SIGKDD International Conference on Knowledge Discovery and Data Mining},
  year={2016}
}

@ARTICLE{Kong2024-tz,
  title     = "{Spatio-Temporal} Pivotal Graph Neural Networks for traffic flow
               forecasting",
  author    = "Kong, Weiyang and Guo, Ziyu and Liu, Yubao",
  journal   = "Proc. Conf. AAAI Artif. Intell.",
  publisher = "Association for the Advancement of Artificial Intelligence
               (AAAI)",
  volume    =  38,
  number    =  8,
  pages     = "8627--8635",
  month     =  mar,
  year      =  2024
}

@article{robinson2021visual,
  title={Visual Analytics for Network Security: A Survey},
  author={Robinson, Connor and others},
  journal={IEEE Transactions on Visualization and Computer Graphics},
  year={2021}
}

@article{chen2023gnn,
  title={Graph Neural Networks for Social Network Analysis: Methods and Applications},
  author={Chen, Jun and others},
  journal={ACM Computing Surveys},
  year={2023}
}

@incollection{kerren2014visual,
  title={Visual analytics: Scope and challenges},
  author={Kerren, Andreas and Stasko, John and Fekete, Jean-Daniel and North, Chris},
  booktitle={Visual Data Mining},
  pages={76--90},
  year={2014},
  publisher={Springer}
}

\end{document}